# PointJEM: Self-supervised Point Cloud Understanding for Reducing Feature Redundancy via Joint Entropy Maximization

Xin Cao[1,2], Huan Xia[1,2], Xinxin Han[1,2], Yifan Wang[1,2], Kang Li,[1,2,3] and Linzhi Su[1,2,4]

[1] *School of Information Science and Technology, Northwest University, Xi'an, Shaanxi 710127, China*
[2] *National and Local Joint Engineering Research Center for Cultural Heritage Digitization, Xi'an, Shaanxi 710127, China*
[3] *likang@nwu.edu.cn*
[4] *sulinzhi029@163.com*

## Abstract

Most deep learning-based point cloud processing methods are supervised and require large scale of labeled data. However, manual labeling of point cloud data is laborious and time-consuming. Self-supervised representation learning can address the aforementioned issue by learning robust and generalized representations from unlabeled datasets. Nevertheless, the embedded features obtained by representation learning usually contain redundant information, and most current methods reduce feature redundancy by linear correlation constraints. In this paper, we propose PointJEM, a self-supervised representation learning method applied to the point cloud field. PointJEM comprises an embedding scheme and a loss function based on joint entropy. The embedding scheme divides the embedding vector into different parts, each part can learn a distinctive feature. To reduce redundant information in the features, PointJEM maximizes the joint entropy between the different parts, thereby rendering the learned feature variables pairwise independent. To validate the effectiveness of our method, we conducted experiments on multiple datasets. The results demonstrate that our method can significantly reduce feature redundancy beyond linear correlation. Furthermore, PointJEM achieves competitive performance in downstream tasks such as classification and segmentation.

## 1. Introduction

Point clouds, which can be collected and stored easily, have a wide range of applications in computer vision [1] and autonomous driving [2,3]. Previous research has mostly focused on supervised learning methods, such as PointNet [4] and DGCNN [5]. These methods can extract semantic features of point clouds well but require a large scale of labeled data to achieve satisfactory accuracy. However, point-by-point annotation of point clouds is a challenging task that requires a lot of time, effort, and resources. Since point clouds are composed of disorganized collections of points. Therefore, the development of 3D visual tasks based on point clouds is limited to a certain extent [6]. At present, the problem can be ideally solved by self-supervised representation learning (SSRL).

SSRL can extract features from a large number of unlabeled data and transfer the learned features to various downstream tasks. Specifically, it can provide a well pre-trained model that can be fine-tuned with a small amount of labeled data to achieve excellent performance [7]. Therefore, SSRL have received a lot of attention and have been extremely successful in the field of two-dimensional (2D) computer vision. For example, the pre-trained model on the large-scale image dataset can be migrated to various downstream tasks, such as image classification [8–10], image segmentation [11,12], image denoising [13]. However, the embedding vectors extracted by SSRL usually contain redundant information, i.e., feature redundancy. Feature redundancy can lead to the degradation of network performance and generalization ability. Therefore, how to reduce feature redundancy has become a key issue for researchers. Niu et al. [10] proposed a novel representation learning method that improve the embedding scheme. The method eliminates feature redundancy and achieves better performance with much less training time and memory usage. Besides 2D field, SSRL has also attracted increasing attention in 3D field. Valsesia et al [14], Wang et al [15], Sauder et al [16] proposed generation-based SSRL methods. Nevertheless, these methods typically require long training times and have degraded performance on downstream tasks. There are also some contrastive learning-based methods [17,18]. These methods use data augmentation on a batch of samples to create two variants of similar samples, and then learn a representation that aligns the similar samples in feature space and distances them from the dissimilar ones [19]. Yet, contrastive learning methods are very susceptible to the problem of model collapse [20]. In detail, the network maps all representation to a constant, also called a trivial solution. Info3D [18] and STRL [17] respectively use memory bank and gradient stopping to avoid trivial solutions. Nevertheless, these techniques make the training process more complicated.

In this paper, inspired by [10], we propose PointJEM, a new self-supervised representation learning method that employs a novel embedding scheme. The embedding vector consists of several segments, each of which corresponds to a feature. By maximizing the joint entropy between segments, our method can reduce feature redundancy beyond linear correlation. PointJEM can learn useful representations as opposed to trivial solutions, without memory banks or gradient stop. To validate the effectiveness of the proposed method, we conducted experiments on multiple datasets. The results show that our method could achieve state-of-the-art performance. Furthermore, we used multiple widely used point cloud networks as feature extractors to demonstrate the generalization of our method.

Briefly, the main contributions of this work are as follows:

- We propose a self-supervised representation learning method based on joint entropy, which is able to reduce feature redundancy beyond linear correlation and avoid the model collapse without additional techniques.

- The proposed method does not require asymmetric design. It is easy to implement and flexible and efficient in the training process.

- Experiments on datasets such as ModelNet40 and ShapeNet demonstrate that our method has excellent self-supervised point cloud learning capabilities, which can significantly improve performance on tasks such as classification and segmentation.

## 2. Related Works

*2.1 Supervised learning methods for point clouds*

The supervised learning methods for point clouds can be divided into two categories: one that processes the point cloud by changing its structure, and another that can directly process the raw point cloud. The methods that change the structure can be divided into two categories: voxel-based methods and projection-based methods. The voxel-based converts the point cloud into a voxel grid and extracts features at multiple scales using convolutional neural network (CNN) [21,22]. However, the performance of the method depends on the resolution of the voxelization. The projection-based methods process the point cloud in a multi-view perspective [23,24]. Point clouds are projected onto planes with different viewpoints, and subsequently CNN is used to extract multi-scale features. These multi-scale features are then fused into a descriptor to achieve higher performance. The methods that can directly process raw point clouds can be divided into three categories: point-based methods, graph-based methods, and CNN-based methods. PointNet[4] was the first point-based network to be developed but could only extract global features, losing local information about the point cloud. PointNet++ [25] improves PointNet by adding hierarchical structure to better capture local information and extract multi-scale features. Instead of extracting point features directly, the graph-based methods [5,26] use edge features to represent the relationship between points and their neighbors. Among the graph-based methods, the most widely applied one is DGCNN [5]. The CNN-based method [27,28] perform convolution operations on the point cloud by designing a new convolution kernel. Additionally, PCT [29] has proposed a transformer-based framework that improve the attention mechanism and shows superior performance. Although the above supervised learning methods have been able to learn the point cloud representations well, the lack of large-scale datasets in the point cloud field has become a bottleneck for these methods.

*2.2 Self-supervised learning methods for point clouds*

The SSRL methods can be classified into three categories according to the pre-text tasks: generation-based method, multiple modal-based method, and context-based methods.

Generation-based SSRL methods can be divided into three categories: point cloud self-reconstruction, point cloud completion, point cloud up-sampling. In recent years, self-reconstruction has become one of the most concerned pre-text tasks in SSRL methods. The method employs an encoder to encode the high-dimensional inputs into low-dimensional embedding vectors and then use a decoder to decode the embedding vectors into the initial dimensions, in this process, the representations of the data will be extracted. For example, FoldingNet [30] proposed a general framework for reconstructing point clouds of arbitrary morphology from 2D grids. The method introduces a novel folding-based decoder that can reconstruct even very fine point cloud objects with only small errors. SO-Net [31] extracts the structural information of a point cloud hierarchically through self-organizing mapping (SOM) and fuses it into a vector to represent the features of the point cloud. GraphTER [32] introduces graph transformation equivariant representations into SSRL method for the first time. The method is able to extract potential associations of graphs under various transformations, allowing end-to-end learning of point cloud representations. Based on this, researchers designed other methods based on self-reconstruction [33,34]. Point cloud completion learning structural and semantic information by filling in the missing parts of existing point clouds. Achlioptas et al [35] combined auto-encoder with point cloud completion, which significantly improved the quality of the generated point clouds and demonstrated excellent generalization capabilities. In addition, OcCo [15] completes the point cloud by restoring the occlusion point. Inspired by BERT [36], Point-BERT [37] samples the point cloud, divides it into multiple local point clouds as tokens, masks a portion of the tokens, and recovers them using transformer.

Point-BERT demonstrates excellent generalization capabilities and superior performance. The learned representations can be well transferred to new tasks and datasets. However, both of the above methods require long training times and larger memory resources.

The purpose of point cloud upsampling is to generate a dense point set that is similar in shape to the input point cloud, while removing noise and ensuring uniform distribution. PU-GAN [38] introduces generative adversarial network (GAN) network to the up-sampling task. The method can extract the spatial distribution of the input data more accurately and generate a denser point cloud. In addition, there are PU-Net [39] and PU-GCN [40] that also studied point cloud up-sampling. However, uniformity and chamfer distance are commonly used as evaluation metrics in point cloud up-sampling, this approach cannot transfer learned representations to downstream tasks. Multiple modal-based method extracts representations by maximizing the consistency of the data across modalities. This method can extract additional features from other modalities' data with greater generalization ability. CrossPoint [41] use the relationship between 2D images and 3D point cloud objects for cross-modal learning. However, the approach introduces additional data, making the training process more complex. Context-based methods usually learn context similarity through contrastive learning that reduces the distance between positive samples and increases the distance between negative samples in feature space. Info3D [18] designed a loss function based on mutual information. PointContrast [42] extracting features by comparing different views of the same point cloud at a point level. DepthContrast [43] uses data in different formats for contrastive learning. STRL [17] learns the spatio-temporal representation of point clouds by comparing different point cloud frames. The methods proposed in previous studies either use complex network structures [42,43] or introduce additional mechanisms [17,18] during the training process. In contrast, the method proposed in this paper is simpler. It reduces feature redundancy by maximizing the joint entropy and makes the samples two-by-two independent. The approach can avoid model collapse and learn meaningful features that show better performance in various downstream tasks.

## 3. Methodology

In this section, we propose PointJEM, a self-supervised point cloud understanding method. The framework is shown in Fig. 1. We initially introduce the feature extraction method and the new embedding scheme for point clouds in Section 3.1. Then, we describe the loss function based on joint entropy in Section 3.2. Finally, a theoretical analysis illustrates why our method is able to avoid model collapse without additional mechanisms in Section 3.3.

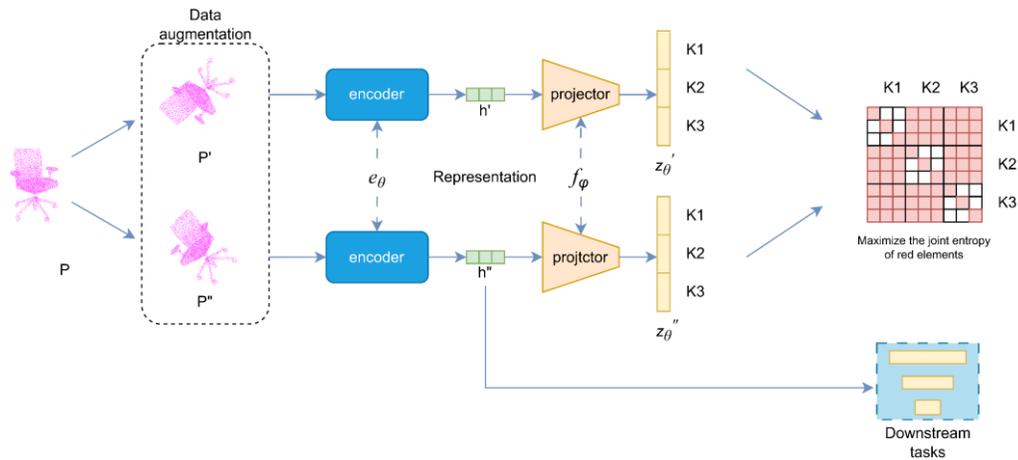

**Fig. 1.** The framework of our method. The input point cloud P are map to two distorted set P′ and P″. The representations vectors $z_\theta{'}$ and $z_\theta{''}$ are extracted by a two-branch encoder $e_\theta$ and projector $f_\varphi$. Network optimization is performed by maximizing the joint entropy of red elements. The representation $h$ obtained by encoders will be used for the downstream tasks.

## 3.1 Preliminaries

The SSRL method consists of two phases: pre-training and downstream tasks. In the pre-training phase, data enhancement techniques such as perturbation, rotation, and scaling are used on a set of point clouds $P = \{p_t\}_{t=1}^{N}$ to obtain two different enhanced versions of the point cloud data $P' = \{p_t'\}_{t=1}^{N}$ and $P'' = \{p_t''\}_{t=1}^{N}$, where N is batch size. The two generated data sets are then fed into two branches, each with an encoder $e_\theta$ and subsequent projector $f_\varphi$, where $\theta$ and $\varphi$ respectively denote the parameters of the encoder and projector to be optimized. The encoder extracts a representation of the point cloud and the projector maps the extracted representation to the embedding space. The two embedded features are denoted as $z_\theta' = f_\varphi(e_\theta(p_t'))$ and $z_\theta'' = f_\varphi(e_\theta(p_t''))$. In the downstream task phase, the projector is discarded and the encoder is retained as the backbone.

Our method employs a new processing method for the obtained embedded feature vectors, dividing them into multiple segments, denoted $z_\theta'(k', m'), k = 1, \cdots, K, m = 1, \cdots, M_k$, where $k$ denotes the number of segments and $M_k$ is the dimension of the $K^{th}$ segment. Each segment represents a type of attribute. We partition the embedding vector uniformly so that each segment has equal dimensions. Then, each segment is converted into one-hot encoding through the softmax function.

$$q_i'(k', m') = \frac{exp(z_\theta'(k', m'))}{\sum_{m=1}^{M_k} exp(z_\theta'(k', m'))} \quad (1)$$

where $q_i'(k', m')$ represents the score of the $m'$-th instance attribute on the $k'$-th segment. The score vectors $q_i''(k'', m'')$ for the other branches are computed in the same way. The attributes represented by each segment will be discrete and complementary after performing the above operations.

## 3.2 Loss base on Joint Entropy

Joint entropy is a measure of uncertainty between a set of random variables. It is often used to calculate the correlation between features. The formula for joint entropy is as follows:

$$H(X, Y) = -\sum_{x \in X} \sum_{y \in Y} P(x, y) \log(P(x, y)) \quad (2)$$

where $x$ and $y$ are specific values of X and Y. Accordingly, $P(x, y)$ is the joint probability of these values occurring together.

In this paper, joint entropy is used to measure the correlation between segments. Before calculating the joint entropy, the joint distribution between every two attributes needs to be estimated. Since each type of attribute has a finite number of discrete instantiations, it is possible to compute the joint probability distribution. Specifically, we can calculate the empirical joint distribution $P(k', k'', m', m'')$ between every two instantiated attributes within and across segments based on a set of samples.

$$P(k', k'', m', m'') = \frac{1}{N} \sum_{i=1}^{N} q_i'(k', m') \, q_i''(k'', m'') \quad (3)$$

where $P(k', k'', m', m'')$ calculates the statistical frequency of a sample having both the m' attribute in k' segment and the m'' attribute in k'' segment over a batch size.

With the empirical joint probability distribution, the information-theoretic measure, i.e., joint entropy, can be directly calculated. The joint entropy is divided into two parts: one for diagonal elements on diagonal blocks and the other for all elements off-diagonal blocks.

The joint entropy of the diagonal block part is calculated as follows:

$$\mathcal{L}_{JED} = \frac{1}{K} \sum_{k',k'',k'=k''} \sum_{m',m'',m'=m''} P(k', k'', m', m'') \log(P(k', k'', m', m'')) \quad (4)$$

The joint entropy of the off-diagonal block part is calculated as follows:

$$\mathcal{L}_{JEO} = \frac{1}{K(K-1)} \sum_{k',k'',k'\neq k''} \sum_{m',m'',m'=m''} P(k', k'', m', m'') \log(P(k', k'', m', m'')) \quad (5)$$

The empirical joint distribution is represented by the block matrix in Fig. 1. The joint entropy of the red part in the matrix will be estimated.

Additionally, we introduce the inner product of two embedded features with different transformations. By optimizing this equation, we can enhance the transformation invariance of the features. Specifically, by minimizing the transform invariance loss, the embedding features of different enhanced versions of the point cloud can be made consistent and encourages the embedding feature within each segment to be one-hot.

$$\mathcal{L}_{ti} = -\frac{1}{NK}\sum_{i=1}^{N}\sum_{k=1}^{K}\sum_{m=1}^{M_k}\log(q_i'(k,m)q_i''(k,m)) \quad (6)$$

In summary, the loss function is defined as follows:

$$\mathcal{L} = \mathcal{L}_{JED} + \mathcal{L}_{JEO} + \lambda\mathcal{L}_{ti} \quad (7)$$

where λ is a balancing factor, usually set to 1, which cannot be too large or too small in order to balance the different parts of the loss function well.

*3.3 Non-trivial solution*

There is usually a problem of model collapse in contrastive learning. Specifically, all samples converge to the same constant solution after mapping into the feature space, producing a trivial solution. To avoid this problem, we need to reduce the correlation between segments so that the attributes learned are different for each segment. In other words, the mutual information between any two segments should be minimized in case the loss function reaches an optimal solution.

$$I(k',k'') = H(k') + H(k'') - H(k',k'') \quad (8)$$

From Eq. (1) and Eq. (3), it follows that the optimal solution to Eq. (4) is: $\forall i, k, m, q_i'(k',m') = q_i''(k'',m'')$, $q_i'(k',:)$ and $q_i''(k'',:)$ are ont-hot vectors, the statistical probability of the $k^{th}$ attribute type taking the $m^{th}$ instantiation is $p(k,m) = \frac{1}{N}\sum_{i=1}^{N} q_i(k,m) = \frac{1}{M^k}$, and $P(k,k,m,m) = \frac{1}{M^k}$. The optimal solution to Eq. (5) is to distribute the samples uniformly over each off-diagonal block. The solution is $\forall k', k'', m', m'', k' \neq k'', P(k',k'',m',m'') = \frac{1}{(M^k)^2}$.

Based on the optimal solution above, we can calculate the value of Eq.8:

$$\begin{aligned}I(k',k'') &= -\sum_{m'=1}^{M^k} p'(k',m')log(p'(k',m')) - \sum_{m''=1}^{M^k} p''(k'',m'')log(p''(k'',m'')) \\ &+ \sum_{m'=1}^{M^k}\sum_{m''=1}^{M^k} P(k',k'',m',m'')log(P(k',k'',m',m'')) \\ &= -log(\frac{1}{M^k}) - log(\frac{1}{M^k}) + log(\frac{1}{(M^k)^2}) = 0\end{aligned} \quad (9)$$

When the mutual information between different segments of the embedding vector is zero, it indicates that the different segments have learned discrete and complementary properties. Therefore, our method can learn informative representations rather than trivial solutions, without requiring asymmetric design.

## 4. IMPLEMENTATION AND EXPERIMENTAL SETUP

*4.1 Dataset*

(1) ModelNet40 [21] is a synthetic dataset that contains 40 categories and 12311 models. 9843 models are used for training and 2468 models are used for testing. Each point cloud contains 2048 points. Following the existing works, we randomly sample 1024 points from each object and augment the data by rotation, translation, etc. (2) ScanObjectNN [44] is a real-world scan dataset that contains 15 categories and 2902 objects. 2321 objects are used for training and 581 objects are used for testing. Due to the background, missing parts and deformations, ScanObjectNN is a highly challenging point cloud classification dataset. Each point cloud contains 2048 points and all the points' coordinates are normalized to the unit sphere. We randomly sample 1024 points from each point cloud. (3) ShapeNet [45] samples from the mesh surfaces of CAD models, and contains 16 categories and 16881 point clouds. 14007 point clouds are used for training and 2874 point clouds are used for testing. Each object is divided into 2-6 parts, with a total of 50 parts. Each sample consists of 2048 points, and the coordinates of all points are normalized. We randomly sample 1024 points from each object. Each point has four attributes: 3D coordinates and category label.

*4.2 Implementation Details*

The pre-trained model uses Adam [46] optimizer, batch size of 48, and epoch of 200. The initial learning rate is set to 1e-3, and the weight decay is 1e-6. The learning rate is gradually reduced by the cosine decay strategy. Following existing works [17], we use rotation, translation and cutout as data augmentation methods. To ensure fair comparison with existing methods, we deploy PointNet and DGCNN as encoders. In addition, we use a two-layer MLP as the projector. The dimensions of the MLP are [embeding,1024]. After the pre-training is finished, the projector is discarded. All downstream tasks are performed on the encoder.

## 5. EXPERIMENTAL EVALUATION

*5.1 3D object class*

### 5.1.1 Linear evaluation for shape classification

Following the standard protocols of STRL [17] and OcCo [15], we evaluate the accuracy of our network model on object classification tasks. We freeze the pre-trained encoder parameters and train a linear classifier. The linear classifier consists of simple MLP layers. Table 1 and Table 2 report the classification results on ModelNet40 [21] and ScanObjectNN [44] respectively. In Table 1, the pre-trained model achieves a competitive accuracy of 90.7% on ModelNet40, surpassing most of the existing methods. "PointNet (ours)" and "DGCNN (ours)" indicate using different backbones for point feature extraction. The results show that the DGCNN backbone has a better feature extraction ability. Our method achieves higher accuracy than most existing methods on ModelNet40. However, it is lower than STRL by 0.2%. Our method performed well on ScanObjectNN and achieved state-of-the-art performance as shown in Table 2.

### 5.1.2 Supervised fine-tuning for shape classification

In addition to evaluating by linear classifiers, we also evaluate our method by semi-supervised learning. The pre-trained model will be used as the initial weights of the feature extractor. We fine-tune on the DGCNN backbone using the ModelNet40 dataset. The results in Table 3 show that our method achieves 0.8% performance improvement compared to "Train from scratch".

Fig. 2. shows the comparison of the learning curves between training from scratch and fine-tuning on ModelNet40. The results show that: (1) Pre-training can speed up the convergence of the model. (2) Fine-tuning can significantly improve the classification accuracy.

Moreover, our method is evaluated under the condition of limited training data being provided. Specifically, the training data will be selected with different ratios, while ensuring that each class has at least one sample. Then, the pre-trained model will be fine-tuned on these limited training data and evaluated on the full test set. Fig. 3 summarizes the results with different ratios of labeled data. Our method can better improve the performance of downstream tasks when fewer training samples are available.

Table 1. Comparisons of the linear evaluation for shape classification on ModelNet40

| Method | Acc |
| --- | --- |
| 3D-GAN [16] | 83.3% |
| Latent-GAN [35] | 85.7% |
| MRTNet [34] | 86.4% |
| SO-Net [31] | 87.3% |
| FoldingNet [30] | 88.4% |
| MAP-VAE [30] | 88.4% |
| 3D-PointCapsNet [30] | 88.9% |
| Sauder et al. + PointNet [47] | 87.3% |
| Sauder et al. + DGCNN [47] | 90.6% |
| STRL + PointNet [17] | 88.3% |
| STRL + DGCNN [17] | 90.9% |
| PointNet (PointJEM) | 88.5% |
| DGCNN (PointJEM) | 90.7% |

Table 2. Comparisons of the linear evaluation for shape classification on ScanObjectNN. our method achieves good performance, which shows that it has good generalization.

| Method | Acc |
| --- | --- |
| PointNet+Jigsaw [47] | 55.2% |
| DGCNN+Jigsaw [47] | 59.5% |
| PointNet+OcCo [15] | 69.5% |
| DGCNN+OcCo [15] | 78.3% |
| PointNet+STRL [17] | 74.2% |
| DGCNN+STRL [17] | 77.9% |
| PointNet+CrossPoint [41] | 75.6% |
| DGCNN+CrossPoint [41] | 81.7% |
| PointNet (PointJEM) | 77.8% |
| DGCNN (PointJEM) | 82.2% |

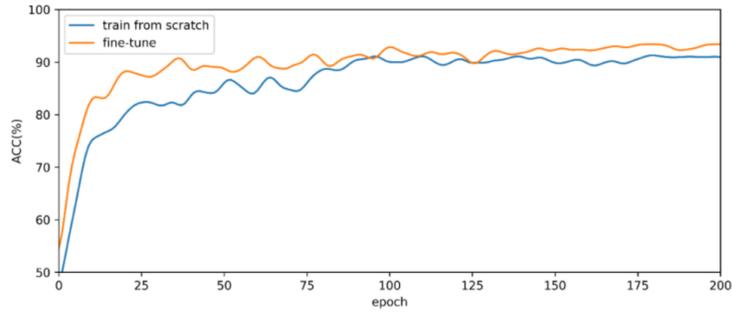

**(a) Accuracy**

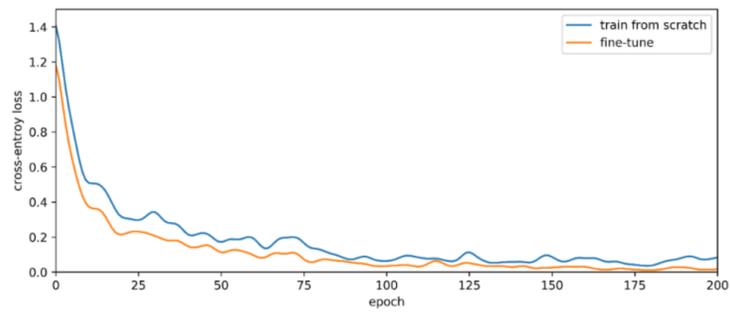

**(b) Cross-entroy loss**

**Fig. 2.** Learning curves in fine-tuning.

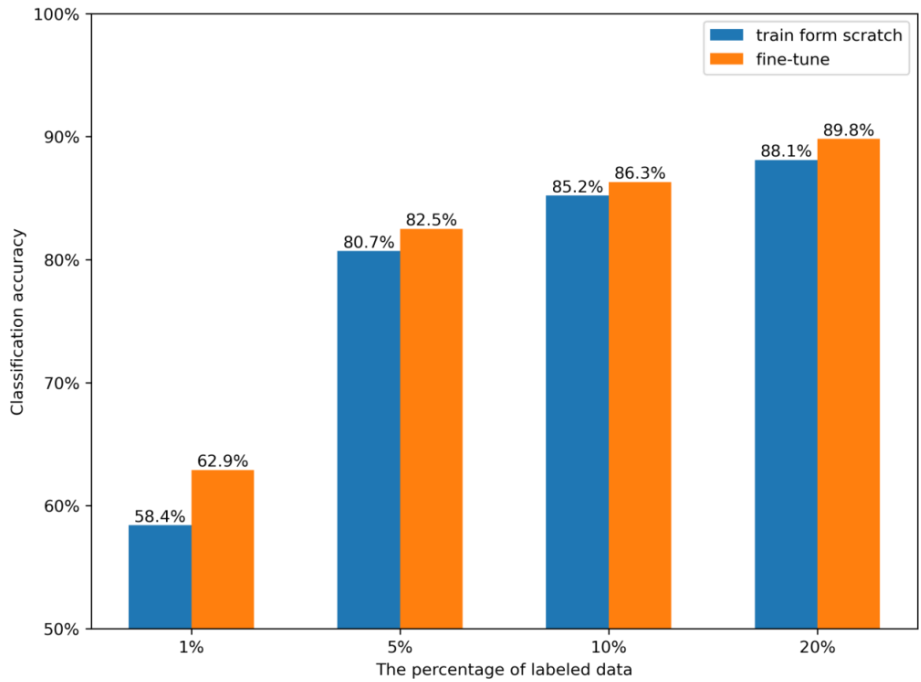

**Fig. 3.** Semi-supervised classification accuracy of different percentages of labeled training data in the ModelNet40 dataset.

Table 3: Comparison of ModelNet40 fine-tune classification results with previous self-supervised methods.

| Method | Acc |
| --- | --- |
| DGCNN+Jigsaw [47] | 92.4% |
| DGCNN+OcCo [15] | 93.0% |
| DGCNN+STRL [17] | 93.1% |
| DGCNN+CrossNet [48] | 93.4% |
| Point-BERT [37] | 93.2% |
| Point-MAE [49] | 93.2% |
| DGCNN (Train from scratch) | 92.2% |
| DGCNN (PointJEM) | 93.0% |

### 5.1.3 Visualization of embeddings

To validate the scalability of the learned embeddings, we use the pre-trained model on ModelNet40 to extract ShapeNet embeddings. As shown in Fig. 4, the ShapeNet embeddings of the ten classes are visualized via T-SNE. It is evident that the distance between intra-class samples is small and the distance between inter-class samples is large. The above results demonstrate the superior generalization of our method.

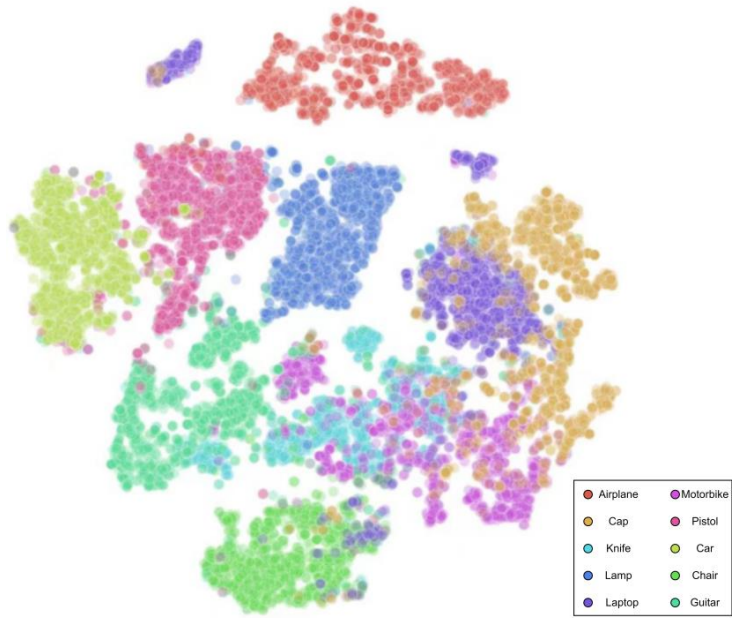

**Fig. 4.** The ShapeNet embedding is visualized in semantic space via T-SNE. The encoder is pre-trained from ModelNet40 and used to extract features from ShapeNet. The results show small intra-class sample distances and large inter-class sample distances, validating the good generalization of our method.

### 5.2 3D Part segmentation

3D object part segmentation is a finer-grained point classification task that requires detailed local features. We directly fine-tune the pre-trained encoder and linear segmenter. mIoU is a widely used evaluation metric for point cloud part segmentation. mIoU is calculated by averaging the IoU of each part in the object, and then averaging the values obtained for each category. Table 4 compares our method with other self-supervised methods. The experimental results show that our method achieves an mIoU value of 85.3%, surpassing the existing two methods. Compared with training from scratch, our method achieves a 0.2% performance improvement. Fig. 5. shows the visualization results of each category in the part segmentation task.

Table 4. 3D object part segmentation fine-tuned on ShapeNetPart

| Method | mIoU |
| --- | --- |
| PointNet [4] | 83.7 |
| PointNet++ [25] | 85.1 |
| PointNet+Jigsaw [47] | 82.2 |
| DGCNN+Jigsaw [47] | 84.3 |
| PointNet+OcCo [15] | 83.4 |
| DGCNN+OcCo [15] | 85.0 |
| Point-BERT [37] | 85.6 |
| Point-MAE [49] | 86.1 |
| DGCNN (Train from scratch) | 85.1 |
| DGCNN (PointJEM) | 85.3 |

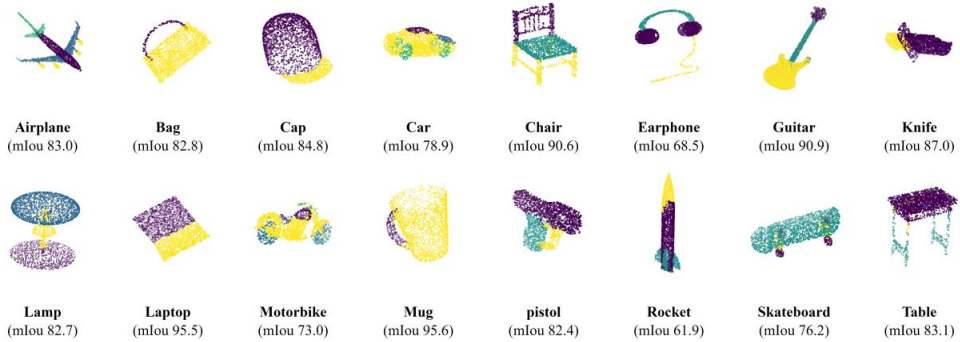

**Fig. 5.** Visualization of 3D Part segmentation results for ShapeNet.

*5.3 Ablation experiment*

5.3.1 Effort of loss function

Table 5 reports the impact of different loss terms on the performance. TI, JED, and JEO respectively represent transformation invariance, diagonal entropy loss, and off-diagonal entropy loss. A performance of 86.8% is achieved with only the joint entropy loss. The result is consistent with the theoretical analysis in 3.3. Without using the JEO loss, the performance drops by 4.8%. The JEO loss helps to learn discrete and complementary attributes in each segment, and also enhances the transformation invariance. In addition, the transformation invariance constraint also significantly improves the performance.

Table 5. Evaluate each part of the loss function

| loss | Acc |
| --- | --- |
| TI | 43.0% |
| TI+JEO | 83.7% |
| TI+JED | 85.2% |
| JED+JEO | 86.8% |
| TI+JED+JEO | 88.5% |

### 5.3.2 Effort of Segment Dimension

The effect of the segment dimension $M_k$ in our method was evaluated. With the same embedding vector dimension, $M_k$ was set to 16, 32, and 64 respectively. Table 6 shows that the best result was obtained when $M_k$ was 32. It also indicates that the performance was not sensitive to this hyperparameter.

Table 6. Evaluate the impact of hyperparameter $M_k$

| $M_k$ | Acc |
|---|---|
| 16 | 88.2% |
| 32 | 88.5% |
| 64 | 87.9% |

### 5.3.3 Effect of Projector Depth

The effect of the MLP layer depth was evaluated. We explored the impact of different MLP layer depths on the performance. Table 7 shows that our method achieved the best result when using 2-layer MLP. It may be due to the correlation between any two segments of the embedding vector is minimized, allowing our method to learn information representations more efficiently.

Table 7. Evaluate the impact of depth of the MLP

| Depth | Acc |
|---|---|
| 2 | 88.5% |
| 3 | 87.8% |
| 4 | 86.4% |

## 6. Conclusion

In this paper, we propose a self-supervised representation learning method that adopts a novel embedding scheme. The scheme divides the embedding vectors into multiple segments, each of which is used to represent a feature. To ensure that the features learned in each segment are discrete and complementary, we design a loss function based on joint entropy. By maximizing the joint entropy between any two segments, the redundancy between segment can be minimized. Our method has a simple structure, is easy to implement, and avoids trivial solutions without requiring asymmetric design. A large number of experiments show that our method has excellent self-supervised representation learning ability and can significantly improve the performance of classification and segmentation downstream tasks. Our approach aims to enhance future point cloud research and help make progress in real-world scenarios with limited annotation. In the future, we plan to explore pre-training techniques for point clouds and devise an efficient method for utilizing large-scale point data.

**Declaration of competing interest**

The authors declare that they have no known competing financial interests or personal relationships that could have appeared to influence the work reported in this paper.

**Data availability**

Data underlying the results presented in this paper are available in Ref. [21,44,45].

## Acknowledgment

This work was supported in part by the National Key Research and Development Program of China (2019YFC1521102, 2019YFC1521103), in part by the Key Research and Development Program of Shaanxi Province (2019GY215, 2021ZDLSF06-04), in part by the National Natural Science Foundation of China (61701403,61806164), and in part by the China Postdoctoral Science Foundation (2018M643719 7).

We would like to acknowledge Dr. Chuang Niu from Rensselaer Polytechnic Institute for providing technical guidance and expertise that greatly assisted our research.